\documentclass[10pt,twocolumn,letterpaper]{article}

\usepackage[pagenumbers]{cvpr}

\usepackage{amsmath,amssymb,booktabs,graphicx}
\usepackage{array,multirow,tabularx,longtable}
\usepackage{microtype}
\usepackage{enumitem}
\usepackage{placeins,needspace}

\usepackage{cuted} 

\newif\ifarxivlayout
\arxivlayouttrue

\newlength{\teaserlift}
\definecolor{cvprblue}{rgb}{0.21,0.49,0.74}
\definecolor{gain}{RGB}{39,111,82}
\definecolor{loss}{RGB}{170,53,55}
\usepackage[pagebackref,breaklinks,colorlinks,allcolors=cvprblue]{hyperref}
\usepackage[capitalize,noabbrev]{cleveref}
\newcommand{\heads}{\mathcal H}

\newcommand{\gain}[1]{\textcolor{gain}{#1}}

\newcommand{\para}[1]{\noindent\textbf{#1}}
\newcolumntype{Y}{>{\raggedright\arraybackslash}X}

\newcommand{\widefig}[5][1.0]{%
  \begin{figure*}[t]
    \centering
    \includegraphics[width=#1\textwidth]{figures/main/#2}
    \caption{#4}
    \label{#3}
  \end{figure*}%
}

\newcounter{finding}
\newcommand{\finding}[1]{\refstepcounter{finding}\noindent\textbf{Finding \thefinding.} \emph{#1}}
\def\paperID{*****}
\def\confName{CVPR}
\def\confYear{2026}
\newcommand{\papertitle}{A Broader Look into MLLM's Retrieval Heads}

\author{%
  Jia Wang$^{1,2*}$ \qquad
  Yuqing Li$^{1,2*}$ \qquad
  Jiangnan Li$^{3*}$ \qquad
  Zheng Lin$^{1,2\dagger}$ \qquad
  Mo Yu$^{3}$\\
  $^{1}$Institute of Information Engineering, Chinese Academy of Sciences\\
  $^{2}$School of Cyber Security, University of Chinese Academy of Sciences\\
  $^{3}$Tencent\\
  {\tt\small \{wangjia,liyuqing,linzheng\}@iie.ac.cn, \{jiangnanli,moyumyu\}@tencent.com}
}

\title{\papertitle}
\begin{document}

\maketitle

\ifarxivlayout
  \preCutedStrip={\vskip-\teaserlift}
\fi

\begin{strip}
\centering
\includegraphics[width=\linewidth]{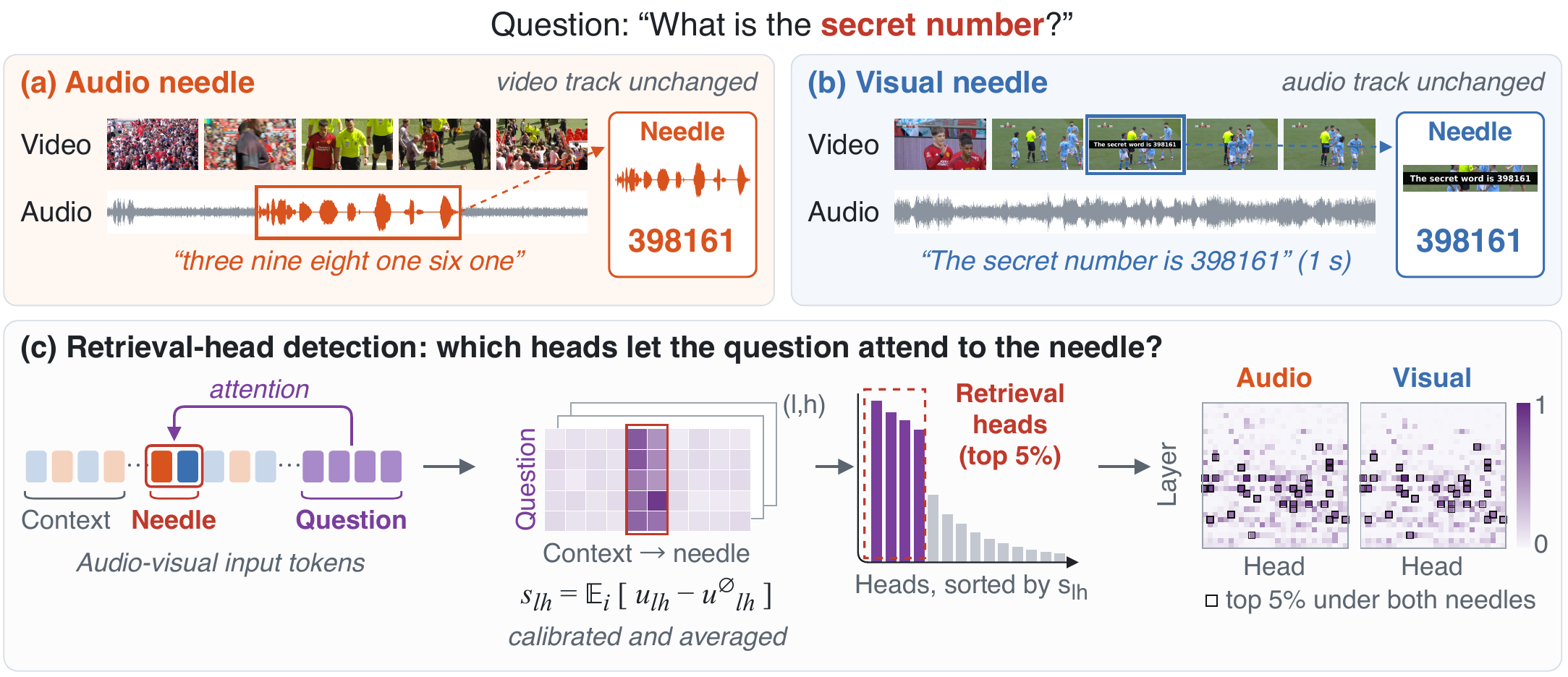}
\captionsetup{hypcap=false}
\captionof{figure}{\textbf{From an AV-NIAH case to retrieval-head scores.} (a,b) Real excerpts of paired AV-NIAH inputs: the same answer is spoken (a) or shown in a one-second banner (b), while the other track is unchanged; boxes mark the needle. (c) We map the needle to input tokens, measure the attention from question to needle in every head, subtract an uninformative-question baseline, and average over examples. The token strip and bars are schematic; the maps show measured Qwen2.5-Omni-7B scores (rows: layers; columns: heads; positive scores normalized per map). Few heads score highly, and most top-5\% heads are shared by the two needles.}
\captionsetup{hypcap=true}
\label{fig:pairing}
\end{strip}

\preCutedStrip={}

{
  \renewcommand{\thefootnote}{\fnsymbol{footnote}}
  \footnotetext[1]{Equal contribution.}
  \footnotetext[2]{Corresponding author.}
}

\begin{abstract}
Locating relevant evidence in long inputs is a core capability of Multimodal Large Language Models. Prior studies attribute this capability to retrieval heads, a sparse set of attention heads that transfer information from the context to the output. However, these analyses are confined to a limited set of modalities and the corresponding models. Moreover, how retrieval heads can be leveraged to enhance the model calls for further exploration. In this paper, we ask whether this mechanism expands together with the input modalities of the model. Taking audio-visual input as a newly added modality, we construct paired inputs in which the same answer is either spoken or shown on screen. With the evidence content fixed, retrieval heads remain sparse in the new modality and are shared to a large extent across modalities. Across 18 language, vision-language, and omni-modal checkpoints, omni-modal models largely preserve the retrieval heads of their backbones. Ablating these heads eliminates nearly all audio-visual retrieval. The heads shared across modalities make up fewer than 2\% of all heads, yet ablating them also degrades document understanding, spatial perception, and visual mathematics. Building on these findings, we turn retrieval heads into a target of optimization by training their attention with a ranking objective. This raises the page Recall@1 of Qwen3-VL-8B on MMDocIR by 6.03 points and improves document retrieval across three model families. Functional attention heads can thus serve both as a lens on multimodal models and as a target for improving them.
\end{abstract}

\section{Introduction}
\label{sec:intro}
Multimodal large language models (MLLMs) increasingly process long, information-rich inputs spanning a growing range of modalities~\cite{mmlongdoc,mlvu}. Within these inputs, the evidence needed to answer a question may occupy only a small part of the context, surrounded by unrelated content. Locating this evidence is therefore central to multimodal understanding. As the range of supported modalities expands, understanding the mechanisms that enable this capability becomes increasingly important.

Research on language models provides a starting point: a small set of attention heads, termed \emph{retrieval heads}, transfers relevant information from the context to the output~\cite{wu2025}. These heads occur across model families, and ablating them sharply reduces long-context retrieval accuracy. Subsequent work develops methods for analyzing retrieval heads~\cite{qrhead} and extends this analysis to vision-language models, where visual retrieval similarly relies on a sparse set of heads~\cite{mmrethead,vrh}.

These findings raise two questions. First, does the same retrieval mechanism extend to new modalities as models expand beyond text and images? This involves examining whether retrieval heads are shared across modalities and retained during adaptation. Beyond explaining model behavior, understanding this mechanism may also guide model improvement. Prior work has focused primarily on analysis, while how to translate these insights into stronger multimodal capabilities remains less explored. We therefore ask a second question: can this mechanism serve as a basis for improving multimodal models?

To address these questions, we first study audio-visual input in omni-modal models. This setting allows the same information to be spoken or displayed, while related language and vision-language backbones allow comparisons across stages of modality adaptation. We construct paired inputs that place the same answer in a long video as either speech or on-screen text, holding the background, question, and answer fixed (\cref{fig:pairing}a,b). Together with text and image-text retrieval tasks, this design supports a unified question-to-evidence analysis across 18 checkpoints. Targeted head ablations then test the functional role of the detected heads. We obtain four findings:
\begin{itemize}[leftmargin=*,itemsep=1pt,topsep=2pt]
\item \textbf{Sparsity (\cref{sec:sparse}).} Retrieval scores concentrate in a small set of heads for both spoken and displayed evidence.
\item \textbf{Cross-modal sharing (\cref{sec:reuse}).} Models largely reuse the same retrieval heads when the same information is presented in different modalities.
\item \textbf{Preservation (\cref{sec:retention}).} Omni-modal models largely retain the retrieval heads of their language or vision-language backbones, and the retained heads remain necessary for retrieval.
\item \textbf{Causal role (\cref{sec:causal}).} Ablating retrieval heads severely disrupts audio-visual retrieval. Ablating the subset shared across modalities, fewer than 2\% of all heads, also substantially impairs performance on broader multimodal reasoning tasks.
\end{itemize}

Together, these findings identify retrieval heads as a compact, shared mechanism for evidence use, motivating direct optimization of their attention. We therefore turn to multimodal document retrieval (\cref{sec:method}), where evidence localization provides a natural ranking signal. We select retrieval heads on document-related tasks and use their question-to-page attention to score candidate pages. A ranking objective trains these scores to favor relevant pages over distractors. On MMDocIR~\cite{mmdocir}, this improves Qwen3-VL-8B's page Recall@1 by 6.03 points over the untrained scorer and also improves layout retrieval without layout supervision, with gains across three model families.

Our contributions are threefold:
\begin{itemize}[leftmargin=*,itemsep=1pt,topsep=2pt]
\item A paired analysis of retrieval heads across evidence modalities and related model checkpoints, characterizing their sparsity, cross-modal sharing, and preservation as input modalities expand.
\item Causal evidence that retrieval heads support audio-visual retrieval and that a small subset shared across modalities also supports broader multimodal reasoning.
\item A ranking method that directly optimizes retrieval-head attention and improves multimodal document retrieval across model families.
\end{itemize}

\section{Related work}
\label{sec:related}

\para{Functional attention heads.}
Early studies identified retrieval heads as a sparse, causally important component of evidence retrieval in long-context language models, using answer-copying behavior to characterize their function~\cite{wu2025}. Query-focused scoring extends this analysis to attention between a query and relevant context, allowing retrieval heads to be identified on realistic question-answering tasks~\cite{qrhead}. Further analyses reveal the dynamic nature of retrieval heads in evidence retrieval~\cite{dynamic}. In vision-language models, functional specialization has been observed in text recognition, spatial grounding, and visual evidence retrieval, with causal interventions linking selected heads to evidence localization and downstream reasoning~\cite{ocrheads,localheads,vrh,core}.

\para{Functional head utilization.}
Despite growing mechanistic understanding, exploiting retrieval heads to improve model capabilities remains relatively underexplored. Existing efforts have focused largely on language models, ranging from attention-based retrieval and efficient inference to interventions that improve evidence use~\cite{qrhead,duoattention,seal,retmask,li2025qrranker}. As mechanistic analysis extends to multimodal models, emerging applications suggest similar potential, but the broader use of these mechanisms to improve multimodal capabilities remains less explored~\cite{mmrethead,vera}. Building on this perspective, we extend the analysis of retrieval mechanisms and explore their potential to improve multimodal evidence use.
\section{Unified retrieval-head detection}
\label{sec:detection}
\subsection{Formulation of Detection Method}
A retrieval head is expected to direct more attention from a question to its evidence than to the rest of the input. Each example consists of a context, a question, and an annotated evidence span. For example $i$, let $Q_i$ denote the question-token positions, $G_i$ the evidence-token positions, and $a^{(i)}_{lh}(q,t)$ the attention from position $q$ to position $t$ in head $h$ of layer $l$. The evidence attention of a head is
\begin{equation}
u^{(i)}_{lh}=\frac{1}{|Q_i|}\sum_{q\in Q_i}\sum_{t\in G_i}a^{(i)}_{lh}(q,t).
\label{eq:attention}
\end{equation}
Part of this attention reflects positional or content preferences that do not depend on the question. Following MMRetHead~\cite{mmrethead}, we therefore recompute it with an uninformative question on the same context, $u^{(i,\varnothing)}_{lh}$, and define the calibrated retrieval score
\begin{equation}
s_{lh}=\frac{1}{N}\sum_{i=1}^{N}\bigl(u^{(i)}_{lh}-u^{(i,\varnothing)}_{lh}\bigr).
\label{eq:score}
\end{equation}
Heads are ranked by $s_{lh}$, as illustrated in \cref{fig:pairing}c. Unless otherwise noted, \emph{retrieval heads} are the top 5\% of heads in this ranking; other head sets are introduced where they are used. All attention is computed under the model's native visibility constraints.

\subsection{Detection data and models}
\label{sec:setup}
We apply this criterion to three families of benchmarks, all of which annotate the evidence needed to answer each question. For text, we use the evidence-annotated subsets of LongMemEval~\cite{longmemeval} and Natural Questions~\cite{nq} processed by QRHead~\cite{qrhead}. For interleaved image-text inputs, we use MM-NIAH~\cite{mmniah}, which places text or natural-image needles in long interleaved contexts.

For audio-visual inputs, we construct paired AV-NIAH examples (\cref{fig:pairing}a,b) following previous work~\citep{avoc} that hold the evidence content fixed and vary only its modality. A six-digit number is inserted into a background video from MLVU~\cite{mlvu} in one of two ways. In the \emph{audio} condition, synthesized speech reads the digits while the video track is unchanged; in the \emph{visual} condition, the number appears in a centered banner for one second while the audio track is unchanged. Both conditions share the source video, the answer, the insertion depth, and the question \emph{``What is the secret number?''} The time span of the speech or banner is mapped to the corresponding audio or video tokens, which form the evidence span $G_i$. 


\begin{table}[t]
\centering
\caption{\textbf{Checkpoints used for mechanism analysis.} Inputs: text (T), image (I), video (V), audio (A). Measurements use the language decoder.}
\label{tab:models-analysis}
\small\setlength{\tabcolsep}{3pt}
\begin{tabularx}{\linewidth}{lYl}\toprule
Group & Checkpoints & Input\\\midrule
Language & Qwen2.5-3B/7B, Qwen3-4B/8B & T\\
Vision-language & Qwen2.5-VL-3B/7B, Qwen3-VL-4B/8B, MiniCPM-V-4.5 & T, I, V\\
Unified & BAGEL-7B & T, I\\
Omni-modal & Qwen2.5-Omni-3B/7B, MiniCPM-o-2.6/4.5, SALMONN 2+ 3B/7B, SALMONN Pro 4B/8B & T, I, A, V\\\bottomrule
\end{tabularx}\end{table}

We analyze 18 checkpoints spanning language~\cite{qwen25,qwen3}, vision-language~\cite{qwen25vl,qwen3vl,minicpm-o26,minicpm-v45}, unified~\cite{bagel}, and omni-modal models~\cite{qwenomni,minicpm-o45,video-salmonn-2-all} (\cref{tab:models-analysis}). Several omni-modal checkpoints are adapted from language or vision-language backbones in the same set, which makes the retention analysis possible. All measurements target the language decoder that produces textual responses, including the Thinker of Qwen2.5-Omni~\cite{qwenomni} and the understanding expert of BAGEL~\cite{bagel}.

\subsection{Comparison and intervention protocol}
\label{sec:protocol}
\para{Comparing head scores.} Two comparisons structure the analysis. Within a model, changing only the evidence modality tests whether heads are shared across modalities; across related checkpoints, holding the text task fixed tests whether heads are retained under modality adaptation. In both cases, Spearman's $\rho$ over all heads measures agreement of the full ranking, and the overlap of top-ranked sets, compared with the overlap expected by chance, measures agreement among the strongest heads.

\para{Testing necessity by ablation.} Attention scores identify candidate heads; ablation tests whether the model relies on them. We ablate a head set $S$ by zeroing the outputs of its heads before the attention output projection. As a control, we ablate the same number of heads in each layer, drawn at random from outside the retrieval heads, with three seeds. Let $t$ be the task score after ablating $S$ and $\bar r$ the mean score after the matched random ablations. The effect of $S$ is its additional drop
\begin{equation}
E(S)=\bar r-t,
\label{eq:effect}
\end{equation}
so a large $E$ means that removing $S$ is far more harmful than removing an equally sized, equally distributed set of other heads. All ablations use examples held out from head selection; significance is assessed with paired cluster-bootstrap tests and Holm correction.

\begin{figure}[t]
\centering
\includegraphics[width=\linewidth]{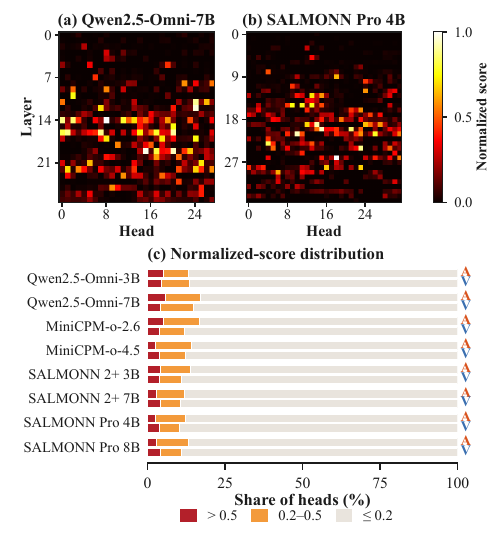}
\caption{\textbf{Retrieval heads are sparse in omni-modal models.} (a,b) Calibrated scores (\cref{eq:score}) under audio evidence for two representative models, normalized to $[0,1]$ per model; rows are layers and columns are heads. (c) Share of heads in three normalized-score bins for eight omni-modal models under audio (A) and visual (V) evidence.}
\label{fig:sparse}
\end{figure}

\section{Retrieval heads in omni-modal models}
\label{sec:analysis}
We examine four properties of retrieval heads as models are extended to audio-visual input: their concentration, their sharing across evidence modalities, their retention from backbones, and their causal role.
\subsection{Retrieval heads are sparse for audio evidence}
\label{sec:sparse}

\finding{Omni-modal models locate spoken and displayed evidence with a small set of retrieval heads.}

\Cref{fig:sparse}a,b shows the score maps of two representative models under audio evidence. Most heads score near zero, and high scores concentrate at a few coordinates in the middle and later-middle layers. The pattern holds in all eight omni-modal models and for both evidence modalities (\cref{fig:sparse}c): fewer than one in ten heads accounts for half of the total positive score in every model. The sparsity reported for text and image evidence~\cite{wu2025,mmrethead} thus also holds when the evidence is speech.

\begin{figure}[t]
\centering
\includegraphics[width=\linewidth]{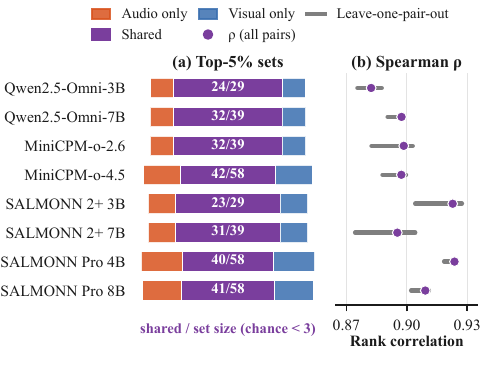}
\caption{\textbf{Spoken and displayed evidence recruit overlapping heads.} (a) Retrieval heads (top 5\%) detected under audio and visual evidence in each model: heads selected in both conditions (purple, with count over set size) and heads selected in only one. Independent random sets would share fewer than three heads. (b) Spearman correlation of all head scores between the two conditions. Gray bars give the range obtained when one of the three detection source--number pairs is left out; they show sensitivity to the detection material rather than a confidence interval.}
\label{fig:overlap}
\end{figure}

\subsection{Audio-visual evidence share retrieval heads}
\label{sec:reuse}

\begin{figure}[t]
\centering
\includegraphics[width=\linewidth]{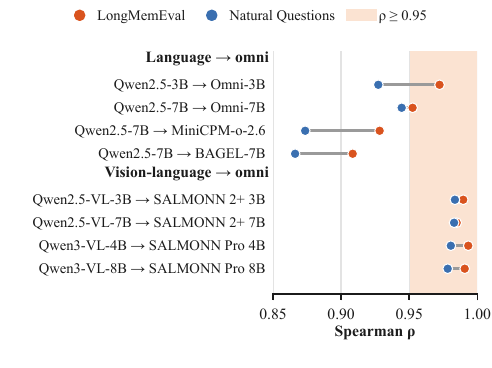}
\caption{\textbf{Retrieval structure is retained across modality adaptation.} Spearman correlation of head scores between a backbone and a related modality-adapted checkpoint on two text retrieval tasks. Shading marks $\rho\geq0.95$.}
\label{fig:adapt}
\end{figure}

\widefig{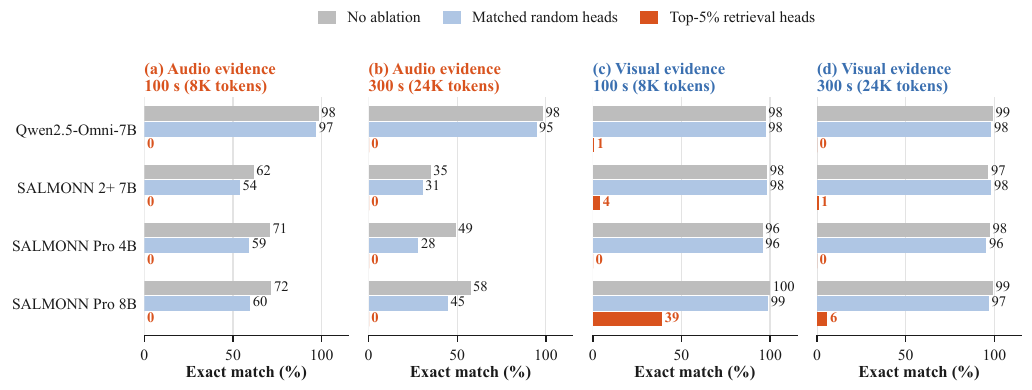}{fig:mask}{\textbf{Retrieval-head ablation disrupts audio-visual retrieval.} Exact match (EM, \%) on held-out AV-NIAH examples without ablation, after ablating an equal number of layer-matched random heads, and after ablating the top-5\% retrieval heads, for audio (a,b) and visual (c,d) evidence with 100-second and 300-second backgrounds (about 8K and 24K input tokens).}{}
\widefig{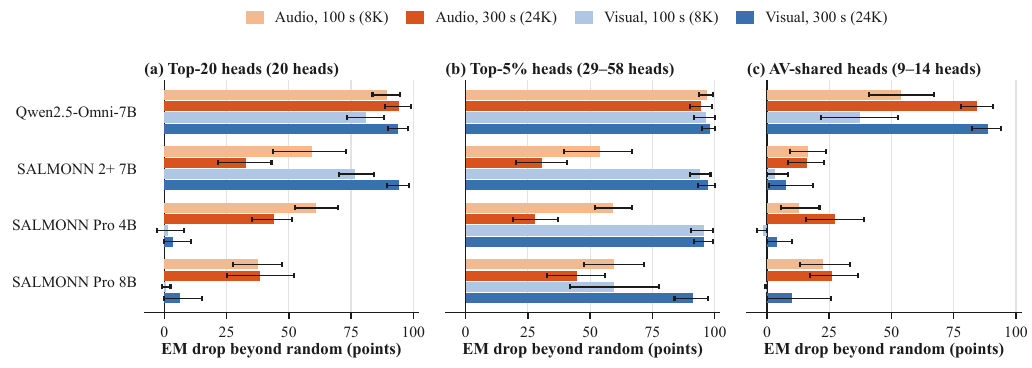}{fig:mask-sets}{\textbf{The ablation effect is robust to how the head set is chosen.} Additional EM drop beyond matched random ablation (\cref{eq:effect}), with 95\% intervals, for three head sets: (a) the top-20 heads of the tested modality, (b) the top-5\% heads, and (c) the AV-shared heads, which appear in both the audio and visual top-20 sets. Bars cover both evidence modalities and background lengths (input tokens in parentheses).}{}
\begin{table*}[t]
\centering
\caption{\textbf{AV-shared heads contribute to downstream multimodal tasks.} Base is the unmodified task score (\%); $E=\bar r-t$ is the additional drop beyond matched random ablation (percentage points; larger means greater damage). AV-shared heads belong to both the audio and visual top-20 sets; Image-ret. heads are selected by natural-image retrieval. Each target set contains 14 heads in Qwen2.5-Omni-7B and 10 in SALMONN Pro 4B, with a separate size- and layer-matched random control. Bold denotes Holm-adjusted $p<0.05$: AV-shared effects are significant on 4/8 tasks per model, versus at most 1/8 for Image-ret.}
\label{tab:transfer}
\small\setlength{\tabcolsep}{6pt}
\begin{tabular}{l r@{\hspace{12pt}}rr r@{\hspace{12pt}}rr}\toprule
&\multicolumn{3}{c}{Qwen2.5-Omni-7B}&\multicolumn{3}{c}{SALMONN Pro 4B}\\
\cmidrule(lr){2-4}\cmidrule(lr){5-7}
& Score (\%) & \multicolumn{2}{c}{Additional drop $E$ (pp)} & Score (\%) & \multicolumn{2}{c}{Additional drop $E$ (pp)}\\
\cmidrule(lr){3-4}\cmidrule(lr){6-7}
Task & Base & AV-shared & Image-ret. & Base & AV-shared & Image-ret.\\\midrule
MMLongBench-Doc & 24.1 & 5.8 & 5.3 & 39.6 & \textbf{10.0} & 2.5 \\
CV-Bench & 75.5 & \textbf{16.3} & 5.7 & 80.5 & \textbf{14.8} & \textbf{6.8} \\
MMStar & 67.0 & \textbf{12.2} & 2.8 & 63.0 & 5.3 & 0.3 \\
CountBenchQA & 83.5 & \textbf{9.6} & 3.4 & 87.4 & 2.2 & $-$5.2 \\
MMMU & 52.5 & 4.0 & 6.2 & 55.5 & 6.7 & 2.2 \\
MMMU-Pro & 39.5 & 8.2 & 4.0 & 39.0 & 3.0 & $-$1.0 \\
MathVista & 72.5 & \textbf{22.3} & 2.5 & 70.5 & \textbf{12.2} & 2.2 \\
MathVision & 29.5 & 8.2 & 3.0 & 49.5 & \textbf{18.8} & 2.3 \\
\bottomrule\end{tabular}\end{table*}

\finding{Matched audio and visual evidence recruit largely the same retrieval heads.}

For each omni-modal model, we compare the paired conditions, which differ only in whether the answer is spoken or displayed. Between 69\% and 83\% of the top-5\% heads are selected under both conditions, whereas independent random sets would share fewer than three heads (\cref{fig:overlap}a). The agreement extends to the full ranking, with $\rho$ between 0.88 and 0.92 in all eight models (\cref{fig:overlap}b), and it persists when any one detection video is left out and with 300-second backgrounds.

The retrieval target accounts for the remaining differences. When spoken digits are compared with natural-image evidence from MM-NIAH, the correlations are lower. The same digit string engages one set of heads whether it is spoken or displayed, whereas a different kind of target engages a partly different set. Head recruitment therefore depends on the retrieval target more than on the evidence modality, which is consistent with the partial sharing observed between text and image retrieval, where both the modality and the content of the evidence differ~\cite{mmrethead}.

\subsection{Retrieval heads are retained along modality adaptation}
\label{sec:retention}

\finding{Omni-modal models largely preserve the retrieval heads of their language or vision-language backbones.}

Vision-language models are known to preserve the retrieval heads of their language backbones~\cite{mmrethead}. Omni-modal models add a further adaptation stage, so we compare checkpoints along the full chain while holding a text retrieval task fixed (\cref{fig:adapt}). Every adapted model remains strongly correlated with its backbone. The SALMONN models are nearly identical to their vision-language backbones ($\rho\geq0.97$), and Qwen2.5-Omni remains close to the language model from which it is adapted. Together with \cref{sec:reuse}, these results indicate that audio-visual adaptation largely reuses the retrieval heads of the backbone and extends them to new evidence modalities.

\subsection{Causal contribution of retrieval heads}
\label{sec:causal}

\finding{Retrieval heads are necessary for audio-visual retrieval, and the heads shared by audio and visual evidence also contribute to broader multimodal tasks.}

\para{Audio-visual retrieval.}
Following \cref{sec:protocol}, we ablate the retrieval heads on held-out examples (\cref{fig:mask}). Removing the top-5\% heads eliminates audio retrieval in all four models and removes most visual retrieval at both background lengths. In Qwen2.5-Omni-7B, for example, audio exact match falls from 98.3\% to 0, whereas matched random ablation leaves it at 96.9\%.

\para{Robustness to the head set.}
The effect does not depend on the particular choice of the top-5\% cutoff (\cref{fig:mask-sets}). The top-20 heads already remove most audio retrieval. The AV-shared heads, about a dozen heads selected under both evidence modalities, also lower audio retrieval well beyond random ablation; in Qwen2.5-Omni-7B, they also lower visual retrieval, and their effect grows with context length.

\para{Transfer to downstream tasks.}
Ablating retrieval heads is known to impair downstream reasoning in vision-language models~\cite{mmrethead,vrh}. We test whether the much smaller cross-modal subset carries this role, using eight benchmarks that span document question answering, visual perception, multidisciplinary reasoning, and visual mathematics (\cref{tab:transfer}). An equally sized set of heads detected on natural-image retrieval serves as a second control. Ablating the AV-shared heads lowers performance beyond random ablation on all tasks, most strongly on visual mathematics and spatial perception, and its effect exceeds that of the image-retrieval heads in 15 of 16 model--task pairs. Cross-modal sharing thus singles out a small set of heads that is central to evidence use well beyond the task on which it was detected.

\section{Retrieval-head-guided document ranking}
\label{sec:method}

\Cref{sec:analysis} shows that a small set of heads carries question-to-evidence attention and that this attention is necessary for using evidence. We now test whether training this attention improves a concrete retrieval task.

\subsection{Task and candidate scoring}
Given a question and candidates $\mathcal D=\{d_1,\ldots,d_m\}$ from one document, the model ranks the candidates by relevance. MMDocIR~\cite{mmdocir} defines two granularities: \emph{Page}, where each candidate is a page screenshot, and \emph{Layout}, where candidates are regions such as paragraphs, tables, or figures. For a fixed retrieval-head set $\heads_k$, the question positions $Q$, and the visual-token positions $T_j$ of candidate $d_j$, we score
\begin{equation}
r_j(\theta)=\sum_{(l,h)\in\heads_k}\frac{1}{|Q|}\sum_{q\in Q}\sum_{t\in T_j}a^\theta_{lh}(q,t),
\label{eq:ranking-score}
\end{equation}
We sum \cref{eq:attention} over selected heads, using each candidate as the evidence span.

\widefig[1.0]{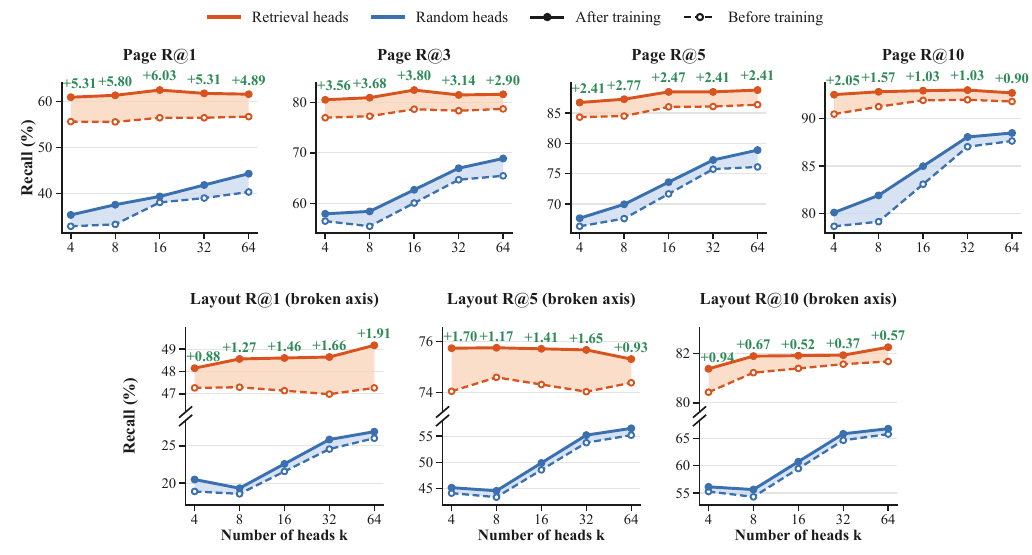}{fig:headcount}{\textbf{Functional head selection matters at every head-set size and cutoff.} Recall of Qwen3-VL-8B when $k$ retrieval heads (orange) or $k$ random heads (blue) define the ranking score, before (dashed) and after (solid) training; shading marks the training gain. Top: Page R@1/3/5/10. Bottom: Layout R@1/5/10.}{}

\subsection{Ranking objective}
Training proceeds in three steps.
\begin{enumerate}[leftmargin=*,itemsep=1pt,topsep=2pt]
\item \textbf{Head selection.} We detect $\heads_k$ once in the base model and keep it fixed during training and inference.
\item \textbf{Score normalization.} Raw attention scores have small and variable ranges. Within each example, we rescale them as
\begin{equation}
z_j=\begin{cases}\gamma\,(r_j-b)/\Delta, & \Delta>\epsilon,\\ r_j, & \Delta\leq\epsilon,\end{cases}
\label{eq:scale}
\end{equation}
with $b=\operatorname{sg}(\min_j r_j)$ and $\Delta=\operatorname{sg}(\max_j r_j-\min_j r_j)$, where $\operatorname{sg}$ stops gradients, $\gamma$ sets the logit scale, and $\epsilon$ guards against degenerate ranges. Detaching the endpoints fixes the logit range at $[0,\gamma]$ while gradients still flow through every $r_j$.
\item \textbf{Ranking loss.} Each training example contains one relevant page $j^+$ and one hard negative from the same document, in random order. We minimize
\begin{equation}
\mathcal L_{\mathrm{rank}}=-\log\frac{\exp(z_{j^+})}{\sum_j\exp(z_j)}.
\label{eq:loss}
\end{equation}
\end{enumerate}
The loss is defined only on the selected heads, but its gradients update LoRA adapters~\cite{lora} on the attention projections. At inference, candidates are ranked by $r_j$ minus the score obtained with an uninformative question, as in \cref{eq:score}.

\subsection{Experimental setup}
\label{sec:method-setup}

\para{Models and heads.} We use Qwen3-VL-8B-Instruct~\cite{qwen3vl} as the main model and repeat the experiments with Qwen2.5-VL-7B-Instruct~\cite{qwen25vl} and Gemma3-12B-it~\cite{gemma3} to test whether the effect carries over to other model families. Following MMRetHead~\cite{mmrethead}, we pool the retrieval heads detected under four multimodal detection conditions and keep the $k$ highest-scoring ones. Unless otherwise stated, we use $k=16$; \cref{fig:headcount} varies this choice.

\para{Training data.} Training examples come from the MMDocIR training set, which provides page-level relevance labels but contains no layout-level data. The model therefore never sees layout supervision, and layout retrieval serves as a test of whether page-level training transfers to a finer granularity. Base and trained variants of each model share preprocessing and candidate windows, so their difference reflects training alone.

\para{Evaluation.} We evaluate both granularities of MMDocIR with Recall@$K$ (R@$K$). Page recall counts a hit when an annotated evidence page is among the top-ranked pages, and layout recall measures the fraction of the annotated evidence area covered by the top-ranked regions. We report Micro averages, which weight questions equally, and Macro averages, which weight the ten document domains equally.

\begin{table}[t]
\centering
\caption{\textbf{MMDocIR Recall@1 (\%).} Upper block: Baseline Results. Lower block: Qwen3-VL-8B attention ranking with $k=16$ heads under one shared configuration, so rows are directly comparable.}
\label{tab:retrieval}
\small\setlength{\tabcolsep}{3.6pt}
\begin{tabular}{lrrrr}\toprule
&\multicolumn{2}{c}{Page}&\multicolumn{2}{c}{Layout}\\
\cmidrule(lr){2-3}\cmidrule(lr){4-5}
Method & Macro & Micro & Macro & Micro\\\midrule
\multicolumn{5}{l}{\emph{Baseline results}}\\
DPR &27.20&26.91&19.24&19.09\\
ColBERT &45.93&45.43&31.57&31.26\\
DSE docmatix &50.41&49.98&29.04&29.55\\
ColPali &55.27&54.58&33.28&33.06\\
Col-Phi3 MMDocIR &57.00&57.10&31.10&31.60\\
\midrule
\multicolumn{5}{l}{\emph{Ours: shared configuration}}\\
Random heads &39.31&38.00&20.32&21.57\\
Retrieval heads &58.81&56.45&46.44&47.14\\
\quad + ranking training &\textbf{63.86}&\textbf{62.48}&\textbf{47.78}&\textbf{48.60}\\
Gain (points)&\gain{+5.05}&\gain{+6.03}&\gain{+1.34}&\gain{+1.46}\\
\bottomrule\end{tabular}\end{table}

\begin{table}[t]
\centering
\caption{\textbf{Ranking training across model families.} MMDocIR Micro R@1 (\%) with retrieval heads before and after page-level training; gains in points. Candidate windows differ across families, so rows compare training effects within a model.}
\label{tab:models}
\small\setlength{\tabcolsep}{2.6pt}
\begin{tabular}{l rrr rrr}\toprule
&\multicolumn{3}{c}{Page R@1}&\multicolumn{3}{c}{Layout R@1}\\
\cmidrule(lr){2-4}\cmidrule(lr){5-7}
Model & Base & Trained & Gain & Base & Trained & Gain\\\midrule
Qwen3-VL-8B & 56.45 & 62.48 & \gain{+6.03} & 47.14 & 48.60 & \gain{+1.46} \\
Qwen2.5-VL-7B & 52.41 & 58.10 & \gain{+5.69} & 31.79 & 32.60 & \gain{+0.81} \\
Gemma3-12B & 44.75 & 46.92 & \gain{+2.17} & 23.88 & 26.04 & \gain{+2.16} \\
\bottomrule\end{tabular}\end{table}

\begin{figure}[t]
\centering
\includegraphics[width=\linewidth]{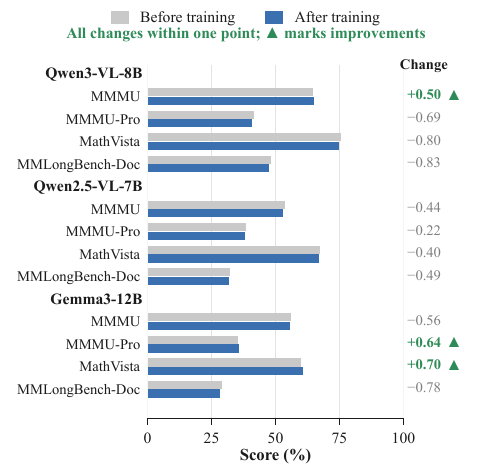}
\caption{\textbf{Ranking training has little effect on out-of-domain performance.} Scores on four out-of-domain benchmarks before and after the main training: MMMU, MMMU-Pro (standard), MathVista, and MMLongBench-Doc. Every change is within one point, and several model--benchmark pairs even improve slightly (green).}
\label{fig:ood}
\end{figure}

\subsection{Results}
\para{Main result.}
\Cref{tab:retrieval} summarizes the results on Qwen3-VL-8B. Even before training, retrieval heads rank candidates far better than random heads, which is consistent with their role in evidence localization. Ranking training then improves page retrieval by 6.03 Micro R@1 points, and the improvement holds in every one of the ten document domains. Layout retrieval also improves, although the training data contain no layout supervision. The trained model even scores above the published retrievers in \cref{tab:retrieval}; because their evaluation settings differ, we treat the gain over the untrained retrieval heads under identical settings as the controlled comparison.

\para{Generality across models and tasks.}
We next apply the same training to the other two model families. The results in \cref{tab:models} show that it improves both granularities for all three models. Page gains are largest for the two Qwen models, and layout retrieval improves for all three. On four out-of-domain benchmarks, scores change by less than one point, with small gains on some model--task pairs (\cref{fig:ood}), indicating the small effect of our training towards out-of-domain performance.

\para{Head-set size.}
Finally, we vary the number of selected heads and compare retrieval heads with equally many random heads (\cref{fig:headcount}). For every head-set size and recall cutoff, training improves the retrieval-head scorer; the largest gains appear in page R@1, where the base model has the most room for improvement. Random heads also benefit from training but remain far below retrieval heads, both before and after training. These results show that functional head selection remains important after ranking training.
\section{Conclusion}
\label{sec:conclusion}

In this paper, we investigated two connected questions: whether evidence-localization mechanisms extend as input modalities expand, and whether understanding these mechanisms can guide model improvement. Using paired audio-visual inputs and a unified analysis criterion across 18 checkpoints, we find that omni-modal models locate spoken and displayed evidence through sparse retrieval heads that are largely shared when evidence content is matched and retained from their backbones. These heads are necessary for audio-visual retrieval, and the small subset shared across modalities also contributes to broader multimodal tasks. This functional role motivates the design of treating retrieval-head attention as a target for optimization. Training selected-head attention with a ranking objective improves multimodal document retrieval across three model families, demonstrating how the mechanism identified through analysis can inform a practical method. Together, these results connect the understanding of multimodal evidence use with its improvement in a broader setting, positioning functional attention heads as both an interpretability lens and an optimization target.

\clearpage
{\small\bibliographystyle{ieeenat_fullname}\bibliography{references}}
\end{document}